\documentclass{amsart}
\usepackage{diagrams}
 \newtheorem{thm}{Theorem}[subsection]

 \newtheorem{prop}[thm]{Proposition}
 \theoremstyle{definition}
 \newtheorem{defn}[thm]{Definition}
 \theoremstyle{remark}
 \newtheorem{rem}[thm]{Remark}

\begin{document}
\title[Monadic Control Constructs for Inference]
 {Monadic Style Control Constructs \\for Inference Systems}

\author{ Jean-Marie Chauvet }

\address{Dassault D\'{e}veloppement, Paris, France}

\email{jmc@neurondata.org}



\date{ August 4, 2002.}

\dedicatory{}


\begin{abstract}
Recent advances in programming languages study and design have
established a standard way of grounding computational systems
representation in category theory. These formal results led to a
better understanding of issues of control and side-effects in functional and imperative
languages. Another benefit is a better way of modelling computational effects in
logical frameworks. With this analogy in mind, we embark on an
investigation of inference systems based on considering inference
behaviour as a form of computation. We delineate a categorical
formalisation of control constructs in inference systems. This
representation emphasises the parallel between the modular articulation of the
categorical building blocks (triples) used to account for the inference
architecture and the modular composition of cognitive processes.
\end{abstract}

\maketitle

\section{Introduction}
Originally grounded in research work in Artificial
Intelligence (AI) and, more specifically in AI in medicine and
production systems \cite{Rappaport_1984_77}, \emph{Nexpert Object}
became a commercial success in the late eighties and
nineties. Its implementation architecture broke new ground in
graphical user interfaces, in portability across operating
environments and in distributed object systems. But powerful
innovations also went deep into the design of the inference system
itself (NXP), which can be traced back to that seminal research
work in AI.

In this and a companion paper we offer a formal account of this
 inference systems, using recent theoretical
advances in the understanding of programming language design.  In
particular, some of the computational effects introduced by the NXP
design show a denotational semantics best captured with the
current tools in \emph{continuation-passing} or
\emph{monadic} styles.

\emph{Continuations.} The idea of transforming programs to
continuation-passing style (CPS) appeared in the mid-sixties
\cite{sabry92reasoning}. The transformation was formally codified
by Fischer and Reynolds in 1972, yielding a standard CPS
representation of call-by-value lambda calculus. In the context of
denotational semantics \cite{Stoy}, described as the theory of
meaning for programs, denotations are usually built with the help
of functions in some mathematical structures. In CPS these
functions are explicitly passed an additional argument $\kappa$
representing "the rest of the computation", a first step towards
full reification of the notion of computation itself. Although
most of the work on continuations involves a functional language,
usually a fragment of typed or untyped lambda calculus, CPS is
also useful in understanding imperative languages. Continuations
were found to be a major tool for the design of interpreters
and compilers for many languages, most prominently Scheme, ML and Haskell.

As such, continuations appear as the raw material of control.
Operations on continuations control of the
unfolding of a computation---that this translation happens in a standard way
is a major result of the theoretical work on continuations.
Similarly, in inference systems, of which production systems are a
well-studied example \cite{PDIS}, designing a proper behaviour
relies on the delicate interweaving of \emph{goal-driven} and
\emph{data-driven} control. Moreover the importance of
the relative data-driven nature of the flow of control sets inference
systems apart from both functional and imperative
programming languages. In inference systems, both side-effects and
procedure calls are mediated through operations on data.
This paper shows the relevance
of the CPS representation for analysis and assessment of inference
systems.

\emph{Monads and triples.} Originally introduced by Moggi \cite{
moggi91notions} in computer science, monads or triples, a notion
from category theory \cite{ttt}, were shown to generalise the
continuation-passing style transformation. Monads can model a wide
variety of features, including continuations, state, exceptions,
input-output, non-determinism, and parallelism \cite{
wadler93monads,wadler94monads}. The \emph{monadic style} essentially
distinguishes between values and computations ; it sets up a
uniform infrastructure for representing and manipulating
computations with effects as first-class objects \cite{
filinski94representing}.  More specifically, monads introduce a
type constructor $T$ through which the type $T \tau$ represents a
computation that yields a result of type $\tau$ and may have
side-effects. Cast in this background, continuations arise as a
special case of monad translation.

\emph{Inference systems.} Inference systems are characterised by a
program organization based on data or event-driven operations.
Such an organisation can be described as \emph{pattern directed}:
patterns occurring in the data select pieces of code in the system
to be activated. Systems organised this way have been used
primarily for deductive inference \cite{PDIS}, and, to a lesser
extent, for inductive inference or learning
\cite{Rappaport_1984_77}. An inference system has three
components: a collection of substructures, called pattern-directed
modules, which can be activated or fired by patterns in the data;
one or more data structures that may be examined and modified by
the pattern-directed modules; and an executive, or interpreter,
that controls the selection and activation of the modules. In
traditional production systems the pattern-directed modules are
called \emph{production rules} and reside in \emph{production
memory} while the data structures are stored in the \emph{working
memory}. The interpreter uses various strategies in a
\emph{recognize-act cycle} to select candidate production rules
 and execute their action part.

In the \emph{NXP architecture}, this mechanism is supplemented by
a goal driven process. Informally, goals are investigated by
actively looking for patterns in data, a search usually called
\emph{backward chaining} as it involves a recursive descent in the
generalized and-or tree representing linked production rules. Such
goals are confirmed, rejected or considered ``not known'' as the
data collected during the recursive descent trigger some of the
patterns of the production rules, a process dually called
\emph{forward chaining}. In addition goals are said to be
\emph{evoked} when, hitherto uninvestigated, they also depend on
data collected during the backward chaining descent from initial
goals. Eventually every goal sharing a triggered data pattern with
an initial goal ends up evoked. These evoked goals are queued for
later evaluation once the current agenda of goal investigation or
data collection completes, their investigation is suspended or
\emph{postponed} until the current one terminates. Considering
backward and forward chainings in NXP as computations in the
previously mentioned sense, goal evocation can be viewed as an
\emph{effect}, comparable, for instance, to the {\tt escape}
construct of functional programming languages. Following sections
of this paper show how to capture backward and forward chainings
as computations and account for goal evocation in
continuation-passing and monadic styles.

\emph{Logical frameworks.} Most of the work we are presenting in
this and its companion paper has been designed for validation and
checking with Twelf \cite{ pfenning99}, a logical framework
developed at Carnegie-Mellon University. In a logical framework
such as this, a deductive system is used as a device for
establishing semantic properties of the mathematical or
computational object under study. Following Martin-L\"{o}f
introduction of \emph{judgements} as the foundation for
constructive mathematics and computer science, logical frameworks
represent judgements as types enabling proof-theoretic derivations
by type reconstruction. Logical frameworks provide a uniform
\emph{judgements-as-types} representation of proofs of semantic
properties. In this paper we use the Twelf deductive system
to investigate the semantic properties of computations with
effects in the NXP inference system (which is itself a deductive
system at a different level of abstraction). Obviously terms such
as \emph{type}, \emph{deduction} and \emph{inference} have
overloaded meanings and we will specify precisely in which context
or system we use the latter when there is a risk of confusion.

The rest of this paper is organised in three sections. The first one
explores a continuation-passing style representation
of the NXP computational architecture. The second presents
generalisation to a  monadic style
description of various effects in the NXP architecture. The last
section uses the monadic results of the previous exploration to
suggest an abstract implementation of the NXP architecture in the
form of the \emph{NXP abstract machine} with implications for
building theoretically sound NXP interpreters or compilers.

\section{Representing inference control with continuations}

\subsection{Terminology}
In the following we will consider simple languages representing
various aspects of inference systems, and, more specifically of
the NXP architecture. Such languages consist of a syntax, defining
the set of well-formed types and well-typed terms, and a
semantics, assigning some notion of meaning to terms. We expect a
semantics to provide a notion of \emph{program evaluation} or
\emph{inference derivation}, usually represented as a (partial)
function from a suitable set of terms to some set of observable
results, say boolean values.

In \emph{denotational semantics} the semantics is expressed as
functions mapping syntactic constructs in the program to
mathematical objects in a mathematical domain. Analyses of
programs are actually conducted in the mathematical domain, in
which properties assessed about denotations of programs translate
to properties about programs. In contrast, the \emph{operational
semantics} defines an abstract machine with a state, possibly
several components, and some set of primitive instructions. The
\emph{axiomatic approach} associates an ``axiom'' with each kind of
statement in the programming language stating what we may
assert after execution of that statement in terms of what was true
beforehand. There is usually a gap between a denotational
semantics and an operational semantics for a language and the main
formal work in designing interpreters and compilers is to prove
their soundness and faithfulness by proving equivalence
between these semantics. Bridging this gap requires to proceed in
several stages: at each stage an alternative semantic definition
of the language is introduces, embodying successively more and more
implementation details. These intermediate stage semantics are
called \emph{non-standard semantics} to reflect their instrumental
status. With these sequence of steps, denotational techniques
often drift into program transformations and non-standard
semantics have often been used as intermediate languages in
compilers.

\subsection{Continuation-passing style transformations}
Practically all programming languages have some form of control
structure or jumping; the more advanced forms of control
structures tend to resemble function calls, so much so they are
rarely describes as jumps. The library function {\tt exit} in C,
for example, may be called with an argument like a function. Its
whole purpose, however, is utterly non functional: it jumps out of
arbitrarily many surrounding blocks and pending function calls.
Such a non-returning function or jump with arguments is an example
of why continuations are needed.

In  continuation-passing style, a function call is transformed
into a jump with arguments to the callee such that one of these is
a continuation that enables the callee to jump back to the caller.
This idea has been formalised into a standard CPS transformation
for lambda calculus.
\begin{defn} \label{d:Lambda-calculus}
The pure untyped lambda calculus $\Lambda$ is defined by \cite{
Barendregt-1997}: A set of terms, $M$, inductively generated over an
infinite set of variables $Vars$,
\begin{itemize}
\item Terms $M ::= V |\: M\: M $; \item Values $V ::= x |\:\lambda
x.M $,  $x$ in $Vars$.
\end{itemize}
\end{defn}

The standards CPS translation for $\Lambda$ in denotational
semantics is given by the map $[[\_]]_{F} : \Lambda \rightarrow
\Lambda$ originally described by Fisher:
\begin{defn} \label{d:FisherCPS}
Let $k$, $m$, $n$ in $Vars$ be variables that do not occur in the
argument to $[[\_]]_{F}$.
\begin{itemize}
\item $[[V]]_{F} = \lambda k.k \psi (V)$ \item $[[M N]]_{F}
= \lambda k.[[M]]_{F}(\lambda m.[[N]]_{F} \lambda n.(m k)n)$ \item
$\psi (x) = x$ \item $\psi (\lambda x.M) = \lambda k.\lambda
x.[[M]]_{F}k$
\end{itemize}
\end{defn}

We will consider a much simpler fragment for the purpose of
analysis of effects and control in inference systems and, more
specifically, in the NXP architecture. The syntax is made up of
simple expressions of boolean arithmetic:
$$ E ::= true \:| false\: | E\: or\: E\: | E\: and\: E $$
In this simplified core model of the inference computation, goals
are represented by and/or tree expressions, the leaves of which are
the working memory elements. Keeping with the essentials,
these working memory elements are valued in ${\bf B} = \{0, 1\}$
the boolean algebra with boolean operations $\&$ (logical and) and
$|$ (logical or). This will be relaxed later as we take into
account other effects in the NXP architecture. In this toy model,
the pattern-directed modules or production rules of inference
systems are captured by expressions in the language, and the
executive system is naturally a boolean arithmetic evaluator.

Following the denotational semantics approach, the standard
semantics for the simple and/or language is defined by:
\begin{defn} \label{d:stdsemNXP}
Let $B = \{true, false\}$ be the set of constants, $Exp$ the set
of expressions inductively defined as in the previous paragraph,
and ${\bf B}$ the classical boolean algebra, we define
the semantic function $\mathcal{B}: B \rightarrow {\bf B}$ and the
evaluation function $[[\_]]_{0}: Exp \rightarrow {\bf B}$ by:
\begin{itemize}
\item $[[B]]_{0} = \mathcal{B}(B)$ \item $[[E_{1}\:or\:E_{2}]]_{0}
= [[E_{1}]]_{0} \:|\: [[E_{2}]]_{0}$ \item
$[[E_{1}\:and\:E_{2}]]_{0} = [[E_{1}]]_{0} \:\&\: [[E_{2}]]_{0}$
\end{itemize}
\end{defn}

This definition does not specify a particular order for evaluation
of boolean expressions, nor does it refer directly to
continuations. As such it is really denotational and only calls
for several steps of refinement to be turned into an operational
semantics. It does, however, reflect the basic principles of
inference systems introduced earlier and, properly extended, will
be an adequate basis on which to build the analysis. The next step
towards operational semantics now consists of providing a non-standard
semantics for the simple language and proving its denotational
equivalence with definition \ref{d:stdsemNXP}.

\subsection{Two non-standard semantics of computations in the NXP architecture}
\subsubsection{Continuations to order goal investigations}
The first non-standard semantics introduced here makes explicit
reference to continuations. In order to better model the behaviour
of inference systems, we add a new operator to our simple and/or
tree language to capture sequential investigation of two or more
goals:
$$ E ::= true \:| false\: | E\: or\: E\: | E\: and\: E | E\:;\:E$$
where $E$ are expressions as in definition \ref{d:stdsemNXP}, and
$E_{1}\:;E_{2}$ describes investigation of goal $E_{1}$
followed by investigation of goal $E_{2}$. In imperative
programming languages, this is simply the succession of
computations of expressions $E_{1}$ and $E_{2}$. We can now define
the following non-standard semantics for this new language as
follows:
\begin{defn} \label{d:nonstdsemNXP1}
Let ${\bf K}$ be the set of continuations, in our case the set of
functions ${\bf B} \rightarrow O$ where $O$ is the type of output
objects with a distinguished element ${\tt exit}: {\bf B}
\rightarrow O$, and $B$, $Exp$ and $\mathcal{B}$ as in definition
\ref{d:stdsemNXP}, we define the evaluation function $[[\_]]_{1}:
Exp \rightarrow {\bf K} \rightarrow {\bf K}$ by:
\begin{itemize}
\item $[[B]]_{1}\:k = k\:\mathcal{B}(B)$ \item
$[[E_{1}\:or\:E_{2}]]_{1}\:k = [[E_{1}]]_{1}(\lambda
\epsilon_{1}.[[E_{2}]]_{1}(\lambda\epsilon_{2}.
k(\epsilon_{1}\:|\:\epsilon_{2})))$ \item
$[[E_{1}\:and\:E_{2}]]_{1}\:k = [[E_{1}]]_{1}(\lambda
\epsilon_{1}.[[E_{2}]]_{1}(\lambda\epsilon_{2}.
k(\epsilon_{1}\:\&\:\epsilon_{2})))$ \item
$[[E_{1}\:;\:E_{2}]]_{1}\:k = [[E_{1}]]_{1}(\:[[E_{2}]]_{1}\:k)$
\end{itemize}
\end{defn}

The continuations introduced here follow the convention of mapping
results of computations, here values of goals inferred by the
system, to a specific output type. With these conventions in
place, the top level {\tt read-eval-print} loop of traditional
interpreters is captured by the simple semantics $$[[E]]_{1}\: {\tt
exit}$$ which basically executes the computation of the expression
$E$ and pass the result to the continuation that yields control
back to the user. In addition, definition \ref{d:nonstdsemNXP1}
introduces a left to right ordering of goal investigation and a
left to right ordering of the data collection process: this
appears in the semantics denotation in places where the left
subexpression of a composed expression is always evaluated first
with a continuation built from the computation of the right
subexpression. This of course, is already a form of design choice
and alternative semantics are easily produced to represent other
options. The embedded design choice then goes one step further
away from the standard denotational semantics down the road of an
implementation-ready operational semantics.

\subsubsection{Sequence semantics for continuations}
With the refined non-standard semantics introduced in this section,
we are setting up a framework for representing the semantics of
control of in the NXP architecture. The following definitions
refine the operational semantics introduced in
\ref{d:nonstdsemNXP1}:
\begin{defn} \label{d:stacks}
We will use the notation $s = \langle B_{1},B_{2}...B_{n}\rangle$
for finite sequences of boolean values, $s$ in ${\bf B}^{*}$ the
set of finite sequences of boolean values; we will also use
indifferently ${\tt nil}$ and $\langle\:\rangle$ to designate the
empty sequence. The finite sequence abstract data type is
characterised by the following operations:
$$s\downarrow i = B_{i} , 0 < i \leq n \;(selection)$$
$$s\uparrow i = \langle B_{i+1},...B_{n}\rangle, 0 < i < n; {\tt nil}, i
= n \;(rest)$$
$$length(s) = n$$
$$s\:+\:s' = \langle
B_{1},B_{2}...B_{n},B'_{1},B'_{2}...B'_{m}\rangle\;(concatenation)$$
$$Or(s) = \langle\;s\downarrow1\:|\:s\downarrow2\;\rangle+s\uparrow2, n \geq 2$$
$$And(s) = \langle\;s\downarrow1\:\&\:s\downarrow2\;\rangle+s\uparrow2, n \geq 2$$
\end{defn}
The sequence abstract data type is comparable to the stack data
type introduced by Stoy \cite{Stoy} for operational semantics of
programming languages. In the NXP architecture the sequence
semantics captures the succession of goals investigated and evoked by
the systems. Based on definition \ref{d:stacks} we are now able to
present the (non-standard) sequence semantics for our simple
representation of the NXP architecture:
\begin{defn}\label{d:nonstdsemNXP2}
Let ${\bf B}^{*}$ be the set of finite boolean values  sequences
with the previously defined abstract data type
operations, let $Exp$ and ${\bf B}$ be as in definition
\ref{d:nonstdsemNXP1} let $\mathcal{B}$ be, as before, the
semantic function of definition \ref{d:stdsemNXP}, we introduce
the new semantic evaluation function $[[\_]]_{2}: Exp \rightarrow
{\bf B}^{*} \rightarrow {\bf B}^{*}$:
\begin{itemize}
\item $[[B]]_{2}s = \langle\mathcal{B}(B)\rangle+s$ \item
$[[E_{1}\:or\:E_{2}]]_{2}\:s =
Or([[E_{2}]]_{2}([[E_{1}]]_{2}\:s))$ \item
$[[E_{1}\:and\:E_{2}]]_{2}\:s =
And([[E_{2}]]_{2}([[E_{1}]]_{2}\:s))$ \item
$[[E_{1}\:;\:E_{2}]]_{2}\:s = [[E_{2}]]_{2}([[E_{1}]]_{2}\:s)$
\end{itemize}
\end{defn}
\begin{rem}\label{r:left-to-right-ordering}
Note that, even though the notation is reversed, compared to
definition \ref{d:nonstdsemNXP1}, the previous evaluation function
still enforces a left to right order of evaluation on expressions
and subexpressions. The left subexpression is always evaluated
first and pushed on the sequence before evaluation of the right
subexpression. This is particularly important in the specific case
of expressions like $E_{1}\:;\:E_{2}$.
\end{rem}
\begin{rem}
In the sequence semantics of the simple NXP language,
continuations are actually ``implemented'' as finite sequences,
initially with type operations similar to stacks (sequence and
rest, for instance). In the NXP architecture, the boolean values
leaves of and/or trees are determined from examination of the
working memory or from interaction with the user (asking
questions). This operation is not captured in the simple language
we have presented so far but will be discussed later. It is important
however it somehow constrains when
operations are actually performed on a finite sequence $s$ of
${\bf B}^{*}$. Namely, we consider that boolean values in a
sequence $s$ are valued \emph{just in time} when they are passed
as arguments to the sequence operations $And$, $Or$, and the like.
(This can be compared to lazy evaluation in languages like Scheme
or Haskell.) Boolean values in sequences of definition
\ref{d:nonstdsemNXP2} are then representations of \emph{delayed}
accesses to the working memory. The reifications of those delayed
executions, usually called \emph{thunks}, are
triggered by the goal investigation process as needed. The lazy
evaluation convention though does not affect the inference computation
nor the control of its unfolding.
\end{rem}
We ground the previous definition with the following result:
\begin{prop} \label{p:congruence}
The  semantics \ref{d:stdsemNXP} and \ref{d:nonstdsemNXP2} are
\emph{congruent} i.e.
$$[[E]]_{2}\:s = \langle[[E]]_{0}\rangle+\:s$$ for all $E$ and for
all $s$.
\end{prop}
\begin{proof}
The proof is by induction on expressions. Obviously :
$$ [[B]]_{2}\:s = \langle\mathcal{B}(B)\rangle +\:s$$ by
definition. For the other syntactic forms of expressions :
\begin{align*}
[[E_{1}\:or\:E_{2}]]_{2}\:s &=
Or([[E_{2}]]_{2}([[E_{1}]]_{2}\:s)) \\
 &= Or([[E_{2}]]_{2}(\langle[[E_{1}]]_{0}\rangle+\:s)) \\
 &=
 Or(\langle[[E_{2}]]_{0}\rangle+\:(\langle[[E_{1}]]_{0}\rangle+\:s))
 \\
 &= Or(\langle[[E_{2}]]_{0},[[E_{1}]]_{0} \rangle+\:s) \\
 &= \langle[[E_{2}]]_{0}\:|\:[[E_{1}]]_{0} \rangle+\:s \\
 &= \langle[[E_{1}\:or\:E_{2}]]_{0}\rangle +\:s
\end{align*}
and similarly for the $and$ expression.
\end{proof}

\subsubsection{Formalisation in a Logical Framework (Twelf)}
The previous approach lends itself to formalisation in a Logical
Framework. In this section, we use the codification of
formalisation techniques into a meta-language offered by Twelf, a
logical framework developed at Carnegie-Mellon University \cite{
pfenning99}, to specify proofs of inference systems properties.
This is done in three stages : The first one is the representation
of the abstract syntax of the simple language under investigation;
the second stage is the representation of the language semantics
(both notions of values and types, and notions of computations,
i.e. operational semantics); the last stage is the representation
of the properties of the language (for instance, congruence of
semantics and type preservation).

We base the representation of the simple language expressions on
abstract syntax (rather than concrete) in order to expose the
essential structure and focus on the semantics and its properties
rather than on lexical analysis and parsing---although the Logical
Framework could also help here. The representation technique is
called \emph{higher-order abstract syntax} and basically uses
types in Twelf to capture all syntax.
\begin{defn} \label{d:TwelfSyntax}
Going back to definition \ref{d:stdsemNXP} of the simple NXP
language, we then declare a new Twelf type: $${\tt exp : type.}$$
to represent expressions of $Exp$, intending that every Twelf
object $M$ of type {\tt exp} represents a simple language
expression and vice-versa. Similarly, we declare boolean values as
a new Twelf type: $${\tt bool : type.}$$ and declare two
distinguished instances of this newly defined {\tt bool} type:
$${\tt true : bool.}$$ $${\tt false : bool.}$$ With these basic
types in place we now turn to boolean values expressions built
from $\&$ and $|$ operators. Twelf captures these
operators as \emph{expression constructors} which translate into
constant functional types:
$${\tt \& : bool \:->\: bool \:->\: bool. }$$
$${\tt | : bool \:->\: bool \:->\: bool. }$$
for which we will use infix mode for better readability.
\end{defn}

In a second stage we introduce a representation of deductions
following the idea of \emph{judgements-as-types}. In Twelf,
deductions are objects and judgements are types, so that proofs in
the semantics space are in fact type reconstructions in Twelf. In
order to do so, we introduce a type family {\tt eval0} indexed by
representations of expressions ($Exp$) and boolean values (${\bf
B}$):
$${\tt eval0 : exp \:->\: bool \:->\: type.}$$
such that we have types such as {\tt eval0 e b} which depend on
objects.

Axioms are simply represented as types such as these,
while semantics operations can be viewed as constructors which,
given deductions of their arguments, yield a deduction of
their result.
\begin{defn} \label{d:TwelfStdSemantics}
For the simple NXP language we introduce:
$${\tt constant : bool \:->\: exp.}$$
$${\tt and : exp \:->\: exp \:->\: exp.}$$
$${\tt or  : exp \:->\: exp \:->\: exp.}$$
and the following representation of their respective evaluations:
$${\tt eval0cst \;:\; eval0\; (constant B)\; B.}$$
$${\tt eval0or  \;:\; eval0\; (E2 \;or\; E1)\; (V1 \;|\; V2) \;<-\; eval0\; E1\; V1\; <- \;eval0\; E2\;
V2.}$$ $${\tt eval0and \;:\; eval0\; (E2\; and\; E1)\; (V1\; \&\;
V2)\; <- \;eval0\; E1\; V1\; <-\; eval0\; E2\; V2.}$$ which
capture the standard semantics of definition \ref{d:stdsemNXP}.
\end{defn}
\begin{rem}
Using Twelf built-in deductive capabilities we can now explicit
deductions such as in the following example:
\begin{verbatim}
eval0 (constant true and constant true or constant false) V.
---------- Solution 1 ----------
 V = (false | true) & false.
 A = eval0_and eval0_cst (eval0_or eval0_cst eval0_cst).
\end{verbatim}
which simply shows the result of the boolean operation as {\tt V}
and the deduction as {\tt A}. Note that the deduction {\tt A} may
be considered as a logical proof of the value {\tt V} or
alternatively as the logical program implementing {\tt eval0}
viewed as a function call.
\end{rem}

The non-standard semantics of the NXP architecture are captured
in the same way, with the introduction of an additional
Twelf type, {\tt bool\_seq} for sequences of boolean values.
\begin{defn} \label{d:TwelfNonStdSem2}
The Twelf representation of the sequence semantics for NXP consists
firstly of an evaluation function which builds a functional
programming style representation of expressions of the simple
language :
\begin{verbatim}
%%% The sequence abstract data type
nil    : bool_seq.
atom   : bool -> bool_seq.

select : bool_seq -> int -> bool.
tail   : bool_seq -> int -> bool_seq.
length : bool_seq -> int.
+      : bool_seq -> bool_seq -> bool_seq. %infix right 10 +.

%%% The functional programming style representation
eval2 : exp -> bool_seq -> bool_seq -> type.

eval2_cst : eval2 (constant B) nil (atom B).
eval2_cst2: eval2 (constant B) S (atom B + S).
eval2_or_s : eval2 (E1 or E2) S (or_s S2)
                <- eval2 E1 S S1 <- eval2 E2 S1 S2.
eval2_and_s : eval2 (E1 and E2) S (and_s S2)
                <- eval2 E1 S S1 <- eval2 E2 S1 S2.
\end{verbatim}
and secondly of an evaluation function that performs the
functionally specified operations on the input sequence :
\begin{verbatim}
eval2a : bool_seq -> bool_seq -> type.

eval2a_atom : eval2a (atom B) (atom B).
eval2a_atom_2 : eval2a (atom B + S) (atom B + S')
         <- eval2a S S'.
eval2a_or_s1 : eval2a (or_s (S'')) (atom(B1 | B2) + S')
           <- eval2a S'' (atom B1 + atom B2 + S)
           <- eval2a S S'.
eval2a_or_s2 : eval2a (or_s (S'')) (atom(B1 | B2))
           <- eval2a S'' (atom B1 + atom B2).
eval2a_or_s_nil : eval2a (or_s (A1 + A2 )) (atom(B1 | B2) )
           <- eval2a A1 (atom B1) <- eval2a A2 (atom B2).
eval2a_and_s1 : eval2a (and_s (S'')) (atom(B1 & B2) + S')
           <- eval2a S'' (atom B1 + atom B2 + S)
           <- eval2a S S'.
eval2a_and_s2 : eval2a (and_s (S'')) (atom(B1 & B2))
           <- eval2a S'' (atom B1 + atom B2).
eval2a_and_s_nil : eval2a (and_s (A1 + A2 )) (atom(B1 & B2) )
           <- eval2a A1 (atom B1) <- eval2a A2 (atom B2).
\end{verbatim}
The Twelf representation of the sequence semantics function is simply
the composition of the previously shown functions :
\begin{verbatim}
eval2b : exp -> bool_seq -> bool_seq -> type.
eval2b_1 : eval2b E S S' <- eval2 E S S'' <- eval2a S'' S'.
\end{verbatim}
\end{defn}
\begin{rem}
There are many ways the non-standard semantics of the NXP
architecture could have been mapped to Twelf's metalanguage.
Definition \ref{d:TwelfNonStdSem2} is in fact more ``operational''
than really needed. It separates a functional programming
equivalent of the expression parsed from the execution of this
representation. The previous definition emphasises the distinction
between \emph{instructions} and a \emph{machine} executing
instructions. The impact of this formal distinction will be
discussed in the implementation section of this paper.
\end{rem}
\begin{rem}
Theorem-proving capabilities of the Twelf environment are also
extremely useful in assessing properties of the simple language.
For instance, as a verification of the congruence of semantics we
can query Twelf as follows :
\begin{verbatim}
%query 1 1
eval2b ((constant true and constant false)
    or constant true
    or constant false) nil
   S.
---------- Solution 1 ----------
S = atom ((false | true) | false & true).
A =
   eval2b_1
      (eval2a_or_s2
          (eval2a_or_s1 eval2a_atom
              (eval2a_atom_2
                  (eval2a_atom_2
            (eval2a_and_s2
                (eval2a_atom_2 eval2a_atom))))))
      (eval2_or_s (eval2_or_s eval2_constant2 eval2_constant2)
          (eval2_and_s eval2_constant2 eval2_constant)).

%query 1 1
eval0 ((constant true and constant false)
    or constant true
    or constant false) V.
---------- Solution 1 ----------
V = (false | true) | false & true.
A = eval0_or
    (eval0_and eval0_cst eval0_cst)
        (eval0_or eval0_cst eval0_cst).

\end{verbatim}
which basically shows deductions of proposition \ref{p:congruence} for
a particular expression.
\end{rem}

Within this framework we are now ready to set forth the semantics
of some of the major inference features of the NXP architecture,
and, with the Logical Framework mapping introduced in this
section, ground some of its behavioural properties into
mathematical logic.

\subsubsection{Sequence semantics for goal evocation}
The sequence semantics previously presented was designed to make
it relatively natural and simple to express goal evocation in the
NXP inference architecture. Informally goal evocation, which is
triggered by the data collection process, \emph{queues} a goal for
postponed investigation, once the current threads of
investigation complete. In turn, we will see that this sequence
semantics is suggestive of an operational semantics closer
to implementation.
\begin{defn} \label{d:nxpsyntax2}
Let us refine the simple NXP language with the following: $Exp$,
expressions: $$ E ::= b\;| E \;or\; E | E \;and\;E | E \;;\;E | b
\;post\;E$$ with $b$ in ${\bf B}$ and where the new expression $b
\;post\;E$ collects data value $b$ and evokes goal $E$ (which, as
an expression, is itself an arbitrarily long sequence of
expressions). The other expressions are as in definition
\ref{d:nonstdsemNXP2}.
\end{defn}
The semantics of goal evocation is captured by extending the
previous definition \ref{d:nonstdsemNXP2} of sequence semantics
with the additional mapping:
\begin{defn}\label{f:nonstdsemNXP3}
Let ${\bf B}^{*}$ be the set of finite boolean values  sequences
with the previously defined selection, rest and concatenation
operations, let $Exp$ and ${\bf B}$ be as in definition
\ref{d:nonstdsemNXP2} let $\mathcal{B}$ be, as before, the
semantic function of definition \ref{d:stdsemNXP}, we extend the
 semantic evaluation function $[[\_]]_{2}: Exp \rightarrow {\bf
B}^{*} \rightarrow {\bf B}^{*}$ as follows:
\begin{itemize}
\item $[[B]]_{2}s = \langle\mathcal{B}(B)\rangle+s$ \item
$[[E_{1}\:or\:E_{2}]]_{2}\:s =
Or([[E_{2}]]_{2}([[E_{1}]]_{2}\:s))$ \item
$[[E_{1}\:and\:E_{2}]]_{2}\:s =
And([[E_{2}]]_{2}([[E_{1}]]_{2}\:s))$ \item
$[[E_{1}\:;\:E_{2}]]_{2}\:s = [[E_{2}]]_{2}([[E_{1}]]_{2}\:s)$
\item $[[B\;post\;E]]_{2}\:s =
\langle\mathcal{B}(B)\rangle+\:s\:+[[E]]_{2}\langle\rangle$
\end{itemize}
\end{defn}
\begin{rem}
The definition of the $post$ semantic function captures the postponed
execution of the goal investigation. It effectively replaces the
current continuation with the same \emph{followed} by the evaluation
of a goal or a sequence of goals. The effect is of completing first the
current goal evaluation process and then pursue with the postponed goals
evoked during that initial process.
\end{rem}
\begin{rem}
Although it is implicit in the definition of sequences, the previous
definition relies on a left-to-right ordering of operation on
sequences as in remark \ref{r:left-to-right-ordering}. If we required
an operational semantics closer to implementation, we would enforce
this by letting sequences hold unevaluated expressions, or thunks, as
well as instructions of a virtual machine, transforming the sequence
abstract data type into a conventional stack abstract data type.
\end{rem}
\begin{rem}
In a refined operational semantics, we would need to be a bit more
specific than the previous definition \ref{f:nonstdsemNXP3}. More
formally, we already mentioned that boolean values in goal
expressions are actually evaluated by accessing the working memory
of the inference system (possibly causing further interaction with
the user). The working memory is represented as an environment
$\rho$, a map from a set of variables $Vars$ to boolean values
{\bf B}. Goal expressions then refer to variables and the
semantic function is expressed in term of $\rho$. The map $\rho$
is partial in the sense that its value is not necessarily defined on all
variables at all times. In the occurrence of one of these unvalued variables,
the goal evaluation process triggers an atomic transaction with
the user, requesting the value which is then assigned in $\rho$. In
programming languages this caching effect is also known as \emph{memoizing} and can be
captured in different ways in an operational semantics. The
postponed evaluation of goals then proceeds in the environment as
resulting from the preceding evaluation cycle. In this, postponed
evaluation differs from re-evaluation that would, in contrast,
proceed from scratch in a new environment $\rho_{0}$ totally
undefined.
\end{rem}
With this semantics we can review some elementary results
formalising the intuitions underlying goal evocation in the NXP
architecture.
\begin{prop}\label{p:commutative}
Sequencing and postponing goals are commutative operations in the
following restricted sense:
$$[[B_{1}\;post\;E_{1}\;;\;B_{2}\;post\;E_{2}]]s =
[[B_{1}\;post\;(E_{2}\;;\;E_{1})\;;\;B_{2}]]s$$
and $$[[B_{1}\;post\;E_{1}\;;\;B_{2}\;post\;E_{2}]]s =
[[B_{1}\;;\;B_{2}\;post;(E_{2}\;;\;E_{1})]]s$$
\end{prop}
\begin{proof}
By simple application of the concatenation operation on sequences.
\end{proof}
This simply states that a sequence of goals can be locally evoked either as a
group or individually without affecting the overall behaviour. In the
accompanying paper, this result will be set in the context of
knowledge acquisition mechanisms concurrently running with the
evaluation process. More specifically, this restricted commutativity
will be cast in the context of the chunking processes described by
Newell in the Soar architecture \cite{Soar}. From an implementation perspective,
proposition \ref{p:commutative} may be used for optimisation of the
execution by preprocessing goal expressions, gathering groups of
evoked goals rather than processing them individually. This would seem
particularly useful when building a lazy interpreter for the NXP
architecture.
\begin{rem}\label{r:evocation-as-exception}
Goal evocation in the NXP architecture is reminiscent of
\emph{exceptions} in functional of imperative programming
languages. In the latter the current continuation is replaced
immediately with another one. The behaviour is then to abandon the
current process and switch to another one. In the former the
continuation is just \emph{added} after the current one; the resulting
behaviour being to successively go through the processes to their
completion. The analogy will be better illustrated in the categorical
investigation of the NXP architecture in the next section.
\end{rem}
Goal evocation as exemplified by the NXP architecture is an
instance of associative cognitive mechanisms. Theses associative
processes are at work in the background of the evaluation process
itself interacting with it by driving the \emph{focus of
attention} of the inference system. While the backward and forward
chaining of  the goal evaluation process could be
described as \emph{strong} or \emph{local} focus of attention,
goal evocation represents a form of \emph{global},
\emph{weaker} forward chaining. More generally if \emph{learning}
processes have traditionally been studied separately from the
performance component of inference systems, in recent resaerch work
emphasis has been put on accounting for interactions between
performance and learning \cite{Soar}. In this perspective
understanding and representing of inference control are critical
as control is the operational juncture between the two subsystems.
Furthermore the representation semantics is of direct service in
specifying the implementation of inference systems.

\section{A categorical investigation of inference control}
\subsection{Motivations}
Category theory has in the past few years become a tool of choice for
to investigate properties of programming languages. Because it
focuses on very high-level properties abstracted from a number of
subdomains of mathematics (domain theory, topology, set theory, etc.) its results are
characterized by a wide range of applicability.

Composition if of the essence of categories. It is only natural
that sequencing and its variations, which are the basic building
blocks of inference systems control constructs---as seen in the
previous section---, find a well-fit mathematical expression in
categorical terms. In this section, we suggest such a categorical
construction for analyses of control in inference systems.
Starting from the original insight of Moggi, we use triples, a
categorical construction with its origins in algebra, to define
effects and controls in the NXP architecture. Following Wadler's lines of
exploration we produce a category to account for goal evaluation
and goal evocation in the NXP architecture. Finally, other results
in \emph{monadic style} representation of computations offer
interesting analogies for the representation of behaviours of
inference systems, particularly in \emph{layering} effects around
the core categorical representation of the architecture. The
categorical study of knowledge acquisition in inference
systems---in its multi-faceted forms---constitutes the major part
of the companion paper covering the NXP architecture.

\subsection{Categorical background}
This section defines the basic notions from category theory that
we need in the formalisation of the NXP architecture in
monadic style. Readers are referred to \cite{ttt} for a
comprehensive presentation of categories and triples. Let $C$ be a
category, we denote by $Obj(C)$ the objects of $C$ and by
$Hom(A,B)$ the set of arrows with source object $A$ and target
object $B$.
\begin{defn} \label{d:functor}
If ${\bf C}$ and ${\bf D}$ are categories, a functor $F:{\bf C}
\rightarrow {\bf D}$ is a map for which:
\begin{itemize}
\item  If $f: A \rightarrow B$ is an arrow of ${\bf C}$, then $Ff:
FA \rightarrow FB$ is an arrow of ${\bf D}$; \item $F(id{_A}) =
id_{FA}$; and \item If $g: A \rightarrow B$, then $F(g \circ f) =
Fg \circ Ff$.
\end{itemize}
\end{defn}
A functor is a morphism of categories, a map which takes objects
to objects, arrows to arrows, and preserves source, target,
identities and composition. More generally $F$ {\em preserves} a
property P that an arrow $f$ may have if $F(f)$ has property P
whenever $f$ has. It {\em reflects} property P if $f$ has the
property whenever $F(f)$ has.

A natural transformation is defined as a "deformation" of one
functor to another.
\begin{defn}
If $F:{\bf C} \rightarrow {\bf D}$ and $G:{\bf C} \rightarrow {\bf
D}$ are two functors, $\lambda : F \rightarrow G$ is a natural
transformation from $F$ to $G$ if $\lambda$ is a collection of
arrows $\lambda C \rightarrow \lambda D$, one for each object $C$
of ${\bf C}$, such that for each arrow $g: C \rightarrow C'$ of
${\bf C}$ the following diagram commutes:

\def\ComponentOne{{\lambda C}}
\def\FuncOne{{Fg}}
\def\FuncTwo{{Gg}}
\def\ComponentTwo{{\lambda C'}}
\begin{diagram}
FC & \rTo^{\ComponentOne} & GC \\
\dTo<\FuncOne & & \dTo>\FuncTwo \\
FC' & \rTo^{\ComponentTwo} & GC'\\
\end{diagram}

The arrows $\lambda C$ are the components of $\lambda$. The
natural transformation $\lambda$ is a {\em natural equivalence} if
each component of $\lambda$ is an isomorphism in ${\bf D}$.
\end{defn}

\subsection{Triples, monads and categories for computations}
Triple or monads are, from one point of view, abstraction of
certain properties of algebraic structures, namely monoids. They
are categorical constructs that originally arose in homotopy
theory and were used in algebraic theory. Moggi
\cite{moggi91notions} was the first to discover the connection
between triples and semantics of effects in programming language
design. Since then the monadic style has pervaded theoretical
research on denotational and operational semantics.
\begin{defn} \label{d:triple}
A triple ${\bf T} = (T, \eta, \mu)$ on a category ${\bf C}$ is an
endofunctor $T: {\bf C} \rightarrow {\bf C}$ together with two
natural transformations $\eta: id{_{{\bf C}}} \rightarrow T$,
$\mu: TT \rightarrow T$ subject to the following commutative
diagrams:

\def\mut{{\mu T}}
\def\tmu{{T \mu}}
\begin{diagram}
TTT & \rTo^\mut & TT \\
\dTo<\tmu & {\rm associativity} & \dTo>{\mu} \\
TT & \rTo^{\mu} & T \\
\end{diagram}

expressing associative identity, and:

\def\etat{{\eta T}}
\def\teta{{T \eta}}
\begin{diagram}
T   & \rTo^\etat    & TT        & \lTo^\teta    & T \\
    & \rdTo<{=}     & \dTo>{\mu}& \ldTo>{=}     &   \\
    &               & T         &               &   \\
\end{diagram}

expressing left and right unitary identities.  The component of
$\mu T$ at an object $X$ is the component of $\mu$ at $TX$, while
the component of $T \mu$ at X is $T(\mu X)$; similar descriptions
apply to $\eta$.
\end{defn}

\begin{rem} \label{d:monad}
There is an alternate way of defining a triple based on a result
due to Manes.

Let ${\bf C}$ be a category with:
\begin{itemize}
\item A function $T : Obj(C) \rightarrow Obj(C)$; \item for each
pair of objects $C$ and $D$, a function $Hom(C,TD) \rightarrow
Hom(TC,TD)$, denoted $f \rightarrow f^{*}$; \item for each object
$C$ of ${\bf C}$ a morphism $\eta C : C \rightarrow TC$;
\end{itemize}
subject to the following conditions:
\begin{itemize}
\item For $f : C \rightarrow TD$, $f = \eta TD \circ f^{*}$; \item
for any object $C$, $(\eta C)^{*} = id_{TC}$; \item  for $f : C
\rightarrow TD$ and $g : D \rightarrow TE$, $(g^{*}\circ f)^{*} =
g^{*} \circ f^{*}$; \end{itemize} is equivalent to a triple on
${\bf C}$.
\end{rem}

The equivalence results from constructions of triples from adjoint
pairs separately discovered by Eilenberg-More and by Kleisli. The
function $ (\_)^{*}: Hom(C,TD) \rightarrow Hom(TC,TD)$ is also
known as the Kleisli star. This alternate definition emphasises
the connection between a triple and a monoid, an algebraic
structure with an associative operation and a unit element. Wadler
suggested a straightforward interpretation of the Kleisli star in
programming language semantics. In this context, the purpose of
the star operation is to combine two computations, where the
second computation may depend on a value yielded by the first
\cite{wadler94monads}. More precisely if $m$ is a computation of
type $T \tau_{1}$ and $k$ a function from values to computations
(such as a continuation) $\tau_{1} \rightarrow T\tau_{2}$, then
$k^{*}(m)$, or $m*k$, is of type $T\tau_{2}$ and represents the
computation that performs computation $m$, applies $k$ to the
value yielded by the computation, and then performs the
computations that results. It binds the result of computation $m$
in computation $k$. Different definitions for the triple $T$ and
the star operation then give rise to different monads to represent
different control operators such as {\tt escape/exit}, {\tt
call/cc}, {\tt prompt/control} or {\tt shift/reset}
\cite{wadler93monads,queinnec93library}.

As noted by Wadler and others, monadic and continuation-passing
styles appear closely related \cite{wadler94monads}. The actual
correspondence, however, is formally quite involved. Filinski has
shown the remarkable result that \emph{any} monadic effect whose
definition is itself expressible in a functional language can be
synthetised from just two constructs: first-class continuations
and a storage cell \cite{filinski94representing,
filinski96controlling}.  This has direct consequences on the
feasibility of various implementations of the NXP architecture as
investigated in the last section of this paper.

\subsection{Monadic style semantics for capturing side-effects}
In most programming languages, evaluation may have implicit
side-effects that are not predicted by the type of the expression.
This is typically the case with goal evocation in the NXP
architecture where collecting data may trigger goals for further,
delayed evaluation. From their first introduction in the world of
programming languages triples, or monads, were precisely used to
distinguish between \emph{values}, and \emph{computations} whose
evaluation may have side-effects. The semantic separation leads to
a stratified style where a pure functional language, for instance,
is used to to express the manipulation of values, and one or
several monadic sublanguages are used to express manipulation of
computations \cite{ wadler94monads}. Layering several monadic
sublanguages is formalised using \emph{triple morphisms} as in
Filinski's \cite{ filinski96controlling}. Interactions between
the core functional language and the monadic extensions are
mediated by the type system which keeps track of the computational
effects and their propagation.

Each side-effect introduces a new type of computation associated
to a particular triple $T$. For instance, triples have been defined for
effects such as exceptions, state, and input/output. Generally
speaking, in monadic style semantics the type $T a$ designates a
computation yielding a value of type $a$ with possible
side-effects. The semantics of these side-effects is captured by
proper definition of the natural transformations of triple
$T$ (its unit and Kleisli star). Formally and following Manes'
construction, the unit of the triple turns a value into a
computation that returns that value without side-effect:
$$\eta : a \rightarrow Ta.$$
The Kleisli star applies a function of type $a \rightarrow Tb$ to
a computation of type $Ta$, basically chaining two computations in
succession. Following Wadler's popular notation---of using $*$ as
an infix operator---the star operation:
$$ (*) : Ta \rightarrow (a \rightarrow Tb) \rightarrow Tb$$
represents application of a continuation to a computation of the
given type. In the following paragraphs we will make the analogy
even more explicit by using the notation:
$$m*\lambda x.k$$
where $m$ and $k$ are expressions and $x$ is a variable. The above
can be read as follows: perform the computation $m$, bind $x$ to
the resulting value and perform computation $k$. In so-called
impure languages, the above notation is similar to:
$${\tt let\:x\:=\:m\:in\:k.}$$
Note that, however, that the latter does not properly
distinguishes pure types (no side-effect) from computation types
(with possible side-effects).

We are now ready to present the monadic style definition of the
previous standard and non-standard evaluators for the NXP
architecture.

\subsection{A monadic style presentation of the NXP architecture}
\subsubsection{The standard evaluation triple} The new
presentation is simply a rework of the semantic evaluation
functions of definition \ref{d:stdsemNXP} in terms of the
appropriate triple, unit, and Kleisli star.
\begin{defn}\label{d:msStdSem}
The standard monadic style semantic of the simple and/or
expression language is captured by the map $eval$:
\begin{align*}
eval &: Exp \rightarrow T {\bf B}  \\
eval(b) &= \eta b \\
eval(E_{1}\;or\;E_{2}) &= eval(E_{1})*\lambda
x.eval(E_{2})*\lambda y.\eta(x\;|\;y) \\
eval(E_{1}\;and\;E_{2}) &= eval(E_{1})*\lambda
x.eval(E_{2})*\lambda y.\eta(x\;\&\;y) \\
\end{align*}
\end{defn}
\begin{rem}
In the definition above lambda abstraction binds less tightly and
the application star binds more tightly, so that we can get rid of
parentheses.
\end{rem}
\begin{rem}
The above evaluation is much more flexible in that is separates
values from computations. Values are evaluated through the unit
semantic function $\eta$ of the triple, while the sequencing of
computations is specified by the application star. The variations
in denotational semantics we explored in the successive
definitions of the previous section are investigated here by
simply changing definitions of unit and application.
\end{rem}

\subsubsection{The non-standard evaluation triple}
In the sequence triple, a computation accepts an initial sequence
and returns a value paired with the final sequence.
\begin{defn}\label{d:msSeqSem}
The sequence triple $T : \tau \rightarrow \tau\times {\bf B}^{*}$
with unit $\eta$ defined by $$\eta(b) = \lambda x.(b, \langle b
\rangle + x)$$ where sequence operations are as in definition
\ref{d:stacks}; and with application star $$(*) : T\tau_{1}
\rightarrow (\tau_{1} \rightarrow T \tau_{2}) \rightarrow
T\tau_{2}$$ such that:
$$
m*k  =\lambda x.{\tt let }\:(a, S_{1})\;=\;m\;x\: {\tt in}\:
      {\tt let }\:(b, S_{2})\;=\;k\;a\;x \:{\tt in }\: (b,
      S_{2})
$$
and the evaluator as in definition \ref{d:msStdSem} captures the
non-standard sequence semantics of the NXP architecture.
\end{defn}
\begin{rem}
The sequence triple is similar to the \emph{state triple}
introduced by Moggi and by Wadler to represent programming
languages with instructions operating on a global state. The
seemingly awkward definition of the application star simply
expresses the intermediary sequences and results of first
computing $m$ and then applying $k$ to the result.
\end{rem}
\begin{rem}
For evaluation purposes, the $eval$ semantic function of the
generic interpreter of definition \ref{d:msStdSem} has $T\;{\bf
B}$ as its domain, meaning that computations with the sequence
triple are of type ${\bf B}\times {\bf B}^{*}$ as expected.
\end{rem}
In order to introduce goal evocation as a last touch to the
sequence triple, we note that goal evocation is \emph{only} a
side-effect and does not interfere with values. This critical
characteristic which was somewhat implicit in the previous
continuation passing style presentation can now be made explicit
in the monadic style. We formalise this separation by defining an
additional semantic function in the sequence triple $T$: $post$ of
type $T\;()$, i.e. $() \rightarrow () \times {\bf B}^{*}$. The
$()$ type signals that the $post$ function is only concerned with
effects on sequences and basically ignores pure values.
\begin{defn}
The evocation function for the sequence triple is defined by:
$$post : T\;()$$
and
$$post\; E = \lambda s.((),s + {\tt inr}(E))$$
where {\tt inr} is the right injection ${\bf B}\times {\bf B}^{*}
\rightarrow {\bf B}^{*}$.
\end{defn}
With the evocation function thus defined, we extend the definition
of the standard interpreter to accommodate the goal evocation
expression in the following final definition of the sequence
triple.
\begin{defn}\label{d:msNonStdSeqSem}
The monadic style sequence semantic of the simple and/or
expression language is captured by the map $eval$ in the sequence
triple $T$:
\begin{align*}
eval &: Exp \rightarrow T {\bf B}  \\
eval(b) &= \eta b \\
eval(E_{1}\;or\;E_{2}) &= eval(E_{1})*\lambda
x.eval(E_{2})*\lambda y.\eta(x\;|\;y) \\
eval(E_{1}\;and\;E_{2}) &= eval(E_{1})*\lambda
x.eval(E_{2})*\lambda y.\eta(x\;\&\;y) \\
eval(b \;post\; E) &= post E * \eta b \\
\end{align*}
\end{defn}
In that definition we simply replaced $\eta b$ with $post \;E *
\eta b$ in the computation. Properties of natural transformations
 assure that the call to $post$ indeed
percolates through higher-level goal evaluations. In the
definition \ref{d:msNonStdSeqSem}, as in the previous section
definitions we insist on a left-to-right evaluation ordering of
sequences.
\begin{rem}
Note that the definition of $post$ could be viewed as a family of
semantic functions indexed by goals, $E$ in the representation. A
better notation in this case would be $post_{E}$ to capture goal
indices. Although equivalent in the monadic style presentation
suggested above for the NXP architecture, this alternative
notation will help formalise the knowledge acquisition part of the
NXP inference architecture.
\end{rem}

\section{Operational semantics and implementation issues}
\subsection{Layering effects in the NXP architecture}
Filinski's and Wadler's work
\cite{filinski96controlling,wadler94monads} have emphasised the
compositional nature of triples in representing computations.
Filinski's results in particular are geared towards presenting a
framework for computational effects which makes it possible to
describe effects in a \emph{modular} way---an idea which permeates
the design of the Haskell functional programming language, to
mention only one. Drawing on the latter framework, we are able to
add \emph{inference effects}, so critical to the NXP architecture
and to inference systems at large, \emph{incrementally}. Following
this approach, the NXP architecture is actually specified by a
sequence of definitional translations, each one of which
``translates away'' one level of inference effects. For example,
we can refine the description of the NXP architecture with goal
evocation and dynamic working memory by specifying it as a
composition of a goal evocation and a working memory translation.
Informally we will also talk of \emph{composing} a goal evocation
and a working memory triple.

The monadic composition relies on the following general idea:
assume we have two triples $T$ and $U$ over a base language $L$,
where $U$ is in a certain sense ``more general'' than $T$. There
is a standard translation of effects, $L^{M}$ into the base
language given their monadic representation. Using this
monadic translation we can give two different translations from
$L^{T}$ to $L$: the original monadic translation for $T$ and a
variant translation using $U$-representations of $T$-effects,
inducing the same evaluation semantics. The reader is referred to
\cite{filinski96controlling} for a formal proof of these
results---interestingly enough the crux of the demonstration relies
on the result that continuations are in a precise sense a
\emph{universal} effect: any definable triple can be simulated by
a continuation triple on a language with only first-class
continuations (a la Scheme) and typed states.

We present here other control constructs of the NXP
architecture and propose candidates for their formal monadic
representation. The working memory, a major element of the NXP
architecture, is detailed by itself in the next section.

\subsubsection{Associative links, context}
In the NXP architecture, the data-triggered goal evocation mechanism
is complemented by second goal-triggered goal evocation process based
on associative links between goals and goals or subgoals.
Associative links are defined by a binary relation: the
\emph{contextual} relation.

\begin{defn}\label{d:nxpsyntax3}
The simple base language for inference systems expression is extended
with the {\tt context} expression. Let us define $Exp$, the set of
expressions: $$ E ::= b\;| E \;or\; E | E \;and\;E | E \;;\;E | b
\;post\;E | E\;context\;E$$ with $b$ in ${\bf B}$ and where the new expression $E
\;context\;E$ evaluates the first expression and evokes the second (which, as
an expression, is itself an arbitrarily long sequence of
expressions). The other expressions are as in definition
\ref{d:nxpsyntax2}.
\end{defn}
\begin{rem}
The goal-triggered goal evocation has lower priority than the
data-triggered one. In the concrete expression
$E_{1}\;{\tt context }\;E_{2}$, any goal evoked by the evaluation of
$E_{1}$ through {\tt post} subexpressions will be evaluated
before actual evaluation of $E_{2}$. Priorities aside {\tt context}
and {\tt post} have the same semantics.
\end{rem}
We reuse the sequence triple to capture the semantics of the
context links in the NXP architecture.
\begin{defn}\label{d:msContextSem}
Associative links semantics in the NXP architecture extend the $eval$
semantic map in the sequence triple as follows:
$$eval(E_{1} \;context\; E_{2}) = post E_{2} * eval(E_{1}) $$
\end{defn}
\begin{rem}
Note that the $post$ primitive is now used twice: in the triple
sequence used to capture goal evocation, and in the second triple
sequence used to capture context links.
\end{rem}

In some respect, associative links are higher level control
constructs in inference systems. Compared to pattern-directed modules
of the canonical architecture, they do not rely on actual patterns in
the working memory elements but rather on the behaviour of the
inference process itself, posting goals for later evaluation when the
inference, seen as computation, reaches some particular goal or
subgoal. Alternatively  the composition
of identical sequence triples could also be viewed as the more general
triple defined by  $T : \tau \rightarrow \tau\times {\bf B}^{*} \times
{\bf B}^{*}$ where the second sequence is in fact used for
contextually linked goals.

\subsubsection{The {\tt reset} action}
Independently of the various actions operating on the working memory,
calling for a special treatment as explained in the next section,
the NXP production rules support a {\tt reset} action on a variable
with the immediate effect of setting its value to ``unknown'' regardless
of its previous assignment.

From the behavioural standpoint, the representation of the computation
of such an expression is simply to \emph{unevaluate} it. When {\tt
reset} operates on a data, the semantics is simply to reevaluate it
should the computation need it at some later time. When {\tt reset}
operates on a goal (an expression in the simple NXP language) however,
the situation is quite different: indeed the reset goal, considered
as data in a later computation, should be reevaluated but what of its
antecedents? And what if further computations do not explicitly need
the reset goal?

In the NXP architecture the effect of the {\tt reset} action
propagates back from the goal to its antecedents but does not affect
posted goals from evocation or contextual links. This propagation only
interferes with the computation process by forcing a reevaluation,
i.e. a new expansion of the reset expression. Its semantics is handled
by the working memory executive which mediates all accesses to values.

\subsection{Working memory and user interaction}
In the semantics definition of the previous sections we
oversimplified an essential element of an inference system, namely
the \emph{working memory}. The working memory is holding the data
structures on which the pattern-directed modules operate. In the
traditional view of inference systems, data is often said to
\emph{enter} or \emph{leave} working memory as it is added or
deleted by the action part of those modules. In the cognitive
sciences framework, the working memory is the representation in
inference systems of the \emph{short-term memory}. Inflow and
outflow of data in working memory capture the essentially local
characteristic of short term memory which can only hold a limited
number of items (``the magic number seven'') and only for a short
period of time.

In most inference systems, and particularly in traditional
production systems where pattern-directed modules are production
rules, the working memory is usually structured as a database of
records. Working memory elements are instances of pre-defined or
user-declared data structures with typed attributes. The
pattern parts of the pattern-directed modules simply express
boolean conditions over these attributes or, in some cases such as in
the OPS series of production system languages \cite{OPS5}, the
presence or absence of particular records.

As such the basic structure is similar to the concept of an
\emph{environment} in denotational semantics. An environment tells what
the identifiers mean in an expression: it says what values the
identifiers denote. In order to capture the level of indirections, it
it usual to introduce $Ide$ a set of identifiers (variable names) and
the domain $U$ of environments as $U : Ide \rightarrow {\bf B}$. With
this the simple NXP language complies with the following definition.
\begin{defn}\label{d:nxpSyntax4}
$Exp$ is now the set of expressions : $$ E ::= x\;| E \;or\; E | E
\;and\;E | E \;;\;E |\; x \;post\;E |\; E \;context\;E $$ with $x$
in $Ide$.
\end{defn}

In the NXP architecture, the working memory additionally plays a
much more dynamic role. In a major departure from the \emph{closed
world} assumption of former inference systems, the working
memory of the NXP architecture is also the locus of interaction
with the external environment, whether data is captured through
interactions with users or from other
applications and systems. Indeed the historical
implementation, \emph{Nexpert Object}, is famous for its
pioneering use of API (Application Programming Interface) and
middleware to integrate inference capabilities to entreprise
applications and information systems in heterogeneous
environments. In order to fulfil this role, the NXP working memory
has its own low-level simplified executive module whose
responsibility is to plan for the acquisition of the value of a
particular data required by the inference process. In
most cases, the plan is a simple look-up of the value for
the requested data in memory. If this fails, either because the
value is unknown yet or because a previous inference operation has
reset that particular data item, the value is sought outside the
NXP system either by asking the user (this in
the form of a popup \emph{question} window) or by sending the
request to the middleware layer to distant databases and
applications (\emph{Nexpert Object} has a wide variety of foreign
request mechanisms ranging from file loading and SQL queries up to
COM or Corba-based dynamic queries to application servers). Beyond
the marshalling of the returned value, its type conversion, the
working memory executive is also able to synchronise the request
for value with the higher-level chaining and goal-driven
processes. It may pre-process group of
requests and use caching in order to optimise usage of costly data
access channels, for instance.

A second dimension of dynamic behaviour that sets the NXP working
memory apart from former production systems architecture is the
evolutive nature of the working memory structure. This
aspect is fully developed in the companion paper, but let us
simply state here that the NXP architecture allows for
pre-processing of the working memory elements based on previous
runs of the inference systems. This pre-processing helps determine
the initial \emph{focus of attention} of the system which, in
turn, assigns priorities in the goal-driven process. This
experience-based \emph{skill acquisition} process
differs from historically studied machine learning processes concerned
with alteration of the collection of pattern-directed modules, or
production memory, e.g. chunking \cite{Soar}.

Operations of the working memory mini-executive are transparent to
the semantic definitions of previous sections. An operational
semantics that would be closer to actual implementation would
replace the simple set of boolean constants, ${\bf B}$, with a
family of semantic functions $get_{B}$, indexed on a set of
variables, representing working memory elements and capturing the
simple semantics of request-response interactions with the external
environment. Those functions could be further refined either in an
asynchronous framework, accounting for message-based communication
channels for instance, or in a lazy evaluation framework in which
the mini-executive is implemented as a lazy interpreter.

The monadic style representation of the semantics of NXP
architecture's working memory follows the layered approach
discussed in the previous subsection. By analogy with Haskell's
I/O monads \cite{HaskellBi98}, we introduce the following
definition.
\begin{defn}\label{d:wmTriple}
The working memory monad $W$ describes a computation as a series of
channels paired with the value returned.
\begin{align*}
{\bf type}\: W\;b =& \langle C_{1},\cdots,C_{n}, b\rangle \\
\eta :& b \rightarrow W b \\
\eta x =& \langle \star_{1},\cdots,\star_{n} , x \rangle \\
(*) :& W b_{1} \rightarrow (b_{1} \rightarrow WM\;b_{2})
\rightarrow W\;b_{2} \\
m*k =& {\tt let}\; \langle csx_{1},\cdots,csx_{n},bx \rangle = m
\;{\tt in}\\
  & {\tt let}\; \langle csy_{1},\cdots,csy_{n},by \rangle
= k\;bx \;{\tt in}\\
  & \langle csx_{1};csy_{1},\cdots,csx_{n};csy_{n}, by\rangle \\
\end{align*}
where channels, $C_{i}$, represent distinct request-response
communication channels; $csx_{i}$ denotes state $x$ of channel
$C_{i}$; $\star_{i}$ denotes the undefined (in the strict sense)
state of channel $C_{i}$ ; $bx$ and $by$ denote boolean values
from {\bf B}, and $csx_{i};csy_{i}$ denotes the state of channel
$C_{i}$ after stepping through state $csx_{i}$ then $csy_{i}$.
\end{defn}
\begin{rem}
The previous definition does not specify the operational semantics of
communication channels. In an operational semantics, channels
would use the state monad which simply captures that requests and
responses flowing through the channel change its state. The
definition also leaves the number of channels unspecified. In the
NXP architecture, at least one of these channels is identified as
the user interaction channel through which question-answer based
interactive transactions can be set up.
\end{rem}
\begin{rem}
This definition also leaves the choice of channel state semantics
open. In particular, different semantics are called for synchronous
and asynchronous communication patterns.
Similarly, definition \ref{d:wmTriple} excludes
interaction between communication channels oversimplifying some of
the capabilities of the mini-executive in the \emph{Nexpert
Object} reference implementation of the NXP architecture.
\end{rem}
The advantage of the monadic style is in this instance twofold.
Firstly, in the layered approach outlined in the previous
paragraphs, the composition of monads adequately reflects the
juncture of the inference processes, chainings and goal-driven, with
interactive processes. Here this articulation is
naturally represented by the composition of the NXP monad from
definition \ref{d:msNonStdSeqSem} with the working memory monad
from definition \ref{d:wmTriple}. This architectural pattern will
be reused in the companion paper to present the composition of
inference with associative and dynamic memory processes.

Secondly, it brings forward the nature of the working memory
mini-executive itself and suggests various operational
implementations. In particular, lazy interpretation could be
useful in the case of continuous data streams channels, for
instance, or for combined management of interacting communication
channels---both situations quite frequent in real-time enterprise
application integration.

\subsection{Operational semantics: an NXP machine}
As a last step towards an operational semantic for the NXP
architecture, we will think of the computation as described by the
operation of a state-based machine: $${\tt until}\: term(\sigma)\:{\tt
do}\: \sigma = step(\sigma)$$
At each stage of the machine's operation the state, $\sigma$, is
modified to a new state as specified by the single step state
transformation function $step$; this is repeated until a terminal
state, recognised by the predicate $term$, is reached. The operation
is captured by a formal definition.
\begin{defn}
With $S$ a set of states, $step : S \rightarrow S$ a function from
states to states, and $term : S \rightarrow {\bf B}$ a function from
states to truth values, we define the general {\tt machine} function :
$${\tt machine}(step, term) = fix(\lambda f
\lambda\sigma.{\tt if}\;term(\sigma)\;{\tt then}\;\sigma\;{\tt
else}\;f\circ step(\sigma))$$
where the expression {\tt if ... then ... else} has the usual meaning
and where $fix$ is the fixed point operator (here acting on $f$).
\end{defn}
\begin{rem}
With this very general definition a machine is simply identified by
the couple of functions $step$ and $term$.
\end{rem}
In order to exhibit an example of an NXP machine, we will
transform the sequences of the previously described non standard
semantics into operational sequences. Here the
denotational semantics naturally suggests an operational
semantics: instead of queuing up denoted values in sequences,
we now queue \emph{instructions} for the NXP machine sequences. Each
expression of the simple language is \emph{compiled} into a
sequence of mixed instructions and arguments, constituting a
program then passed to an abstract machine (with proper $term$ and
$step$ definitions).
\begin{defn}\label{d:programs}
We introduce the syntactic domains: ${\bf B}$ of boolean values, $Ins$
of instructions, $Prg = Ins^{*}$ of finite sequences of instructions,
or programs.

The syntax for instructions is:
$$ I ::= {\tt get}\; x\;| {\tt and}\; | {\tt or}\; | {\tt reset}\; x$$
The semantics of this machine code will basically arrange that, as
might be expected, {\tt get} accesses the working memory to
retrieve the value of a particular identifier---possibly triggering
side-effects in the working memory executive---and place it on top
of the sequence; {\tt or} will form the logical or of the top two
elements on the sequence; similarly for {\tt and}; and {\tt reset}
which causes the working memory executive to reset values used
in the evaluation of a variable--possibly triggering side-effects.
\end{defn}
\begin{rem}
With this definition in place we have shifted emphasis from a semantic
function mapping expressions to denotations to a compiling function
mapping expressions to syntactic values, namely sequences of instructions.
\end{rem}
The compiling function transforms expression into programs fit for the
NXP machine.
\begin{defn}\label{d:compiler}
Following the lines of the non standard semantics, we introduce the
compiling function $C : Exp \times Prg \times Prg \rightarrow Prg \times Prg$.
\begin{align*}
C(x,s,p) &= s+\langle {\tt get}\;x \rangle, p \\
C(E_{1}\;or\;E_{2},s,p) &= {\tt let}\; s', p' =
C(E_{2},C(E_{1},s,p))\;{\tt in}\; s'+\langle {\tt or}\rangle, p' \\
C(E_{1}\;and\;E_{2},s,p) &= {\tt let}\; s', p' =
C(E_{2},C(E_{1},s,p))\;{\tt in}\; s'+\langle {\tt and}\rangle, p' \\
C(E_{1}\;;\;E_{2},s,p) &= C(E_{2},C(E_{1},s,p)) \\
C(x\;post\;E,s,p) &= {\tt let}\; s', p' = C(E,\langle\rangle,\langle\rangle) \;{\tt
in}\; s'+\langle {\tt get }\;x \rangle, p' + p \\
C( E_{1}\;context\;E_{2},s,p) &= {\tt let}\; s', p' =
C(E_{2},\langle\rangle,\langle\rangle) \;{\tt in}\;
C(E_{1},s,p),p'+s'+p \\
\end{align*}
\end{defn}
\begin{rem}
In this definition we need two operational sequences $s$ and $p$. The
first one represents direct continuations as processed by
evaluation and evocation of goals. The second one is used to
process contextual links and goal associations.
\end{rem}
\begin{rem}
As remarked in the previous section, the  compilation
process could be captured by a specific triple then blurring the formal
distinction between compilation and semantic  evaluatiation of NXP simple
language expressions.
\end{rem}
We now complete the definition of an operational semantics for the NXP
architecture by displaying a $term$ and a $step$ function for the
compiled code definitions \ref{d:compiler} and \ref{d:programs}.
\begin{defn}
Let $S =  Prg \times {\bf N} \times {\bf B}^{*}$ be a set of
states, and $\sigma = (\Pi,n,s)$ be a typical member of $S$.
Simply enough, we specify the effect of any single instruction on
the stack by defining a function $I: Ins \rightarrow {\bf B}^{*}
\rightarrow {\bf B}^{*}$ as follows:
\begin{align*}
 I( {\tt get}\;x, s)  = &\: \langle get(x) \rangle + s   \\
 I( {\tt reset}\;x, s)  = &\: s   \\
 I( {\tt or}\;,s)  = &\: Or(s)  \\
 I( {\tt and}\;,s)  = &\: And(s)
\end{align*}
\end{defn}
\begin{rem}
In this definition $get$ and $reset$ are invocations of the working
memory executive. In fact $reset: Ide \rightarrow \langle\rangle$
always returns the empty stack while $get: Ide \rightarrow {\bf B}$
returns the boolean value denoted by identifier $x$---both calls having
possible side-effects. The semantics of these calls is represented by the
working memory triple.
\end{rem}
Finally we define an instance of the abstract machine.
\begin{defn}
The NXP virtual machine is defined by the set $S$ of states, and the
$term$ and $step$ functions:
$$step(\Pi,n,s) = (\Pi,n+1,I(\Pi\downarrow i,s))$$
and
$$term(\Pi,n,s) = n \geq length(\Pi)$$
\end{defn}
\begin{rem}
Thus when the machine executes a program $\Pi$ with the
starting stack $s$--usually the empty stack $\langle\rangle$--the
final stack is obtained as specified by the semantic function $M$,
defined by:
\begin{align*}
M & : Prg \rightarrow {\bf B}^{*} \rightarrow {\bf B}^{*} \\
M(\Pi,s) &= in_{3}({\tt machine}(step,term)(\Pi,1,s)) \\
\end{align*}
where $in_{3}$ is the canonical projection of the third element of the
triplet.
\end{rem}
\begin{rem}
Note that the final program passed to the NXP machine is the
concatenation of the two compiled operational sequences, accounting
for the lower priority of the associative links. This is captured in
the following congruence proposition.
\end{rem}
\begin{prop}\label{p:congruenceCompiler}
The effect of running a compiled expression is the one given by the
sequence semantics, more specifically:
$$M(\Pi,s) = [[E]]_{2}s$$
where
$$\Pi = \;{\tt let}\; s',p' = C(E,s,\langle\rangle)\;{\tt in}\:p'+s'$$ for
all $E$, $s$.
\end{prop}
Interestingly enough the compiler is not the only operational
semantics suggested by the non standard semantics studied
above. Building an interpreter for the NXP architecture is also
possible along the same lines as the compiler, but instructions are
then executed on the fly. More generally several implementations can still
be chosen at this stage, somehow independently of the operational
semantics: eager or lazy evaluation, for instance, is applicable to
both operational semantics.

\section{Conclusions and further research}
Inference may be considered as a special form of computing. This
basic tenet of the whole Artificial Intelligence theoretical field
has been comprehensively captured by Newell's \emph{knowledge
level} hypothesis \cite{Newell93}, and if the distinction between
behaviour and beliefs is well established, the nature of behaviour
as computation is still much debated.

Inspired by the analogy between the latter distinction and the one usually
drawn between computations and values in the study and design of
programming languages, we have presented an architectural view of
control constructs in inference systems and, more specifically, of
\emph{Nexpert Object} as a canonical inference system. The resulting
NXP architecture has grounded semantics expressed in categorical terms
using the monadic style of presentation.

The benefits of this formal representation is twofold. On the
one hand, the categorical foundation of this representation helps
study of the NXP architecture in a logical framework, such as Twelf,
and formally assess its architectural properties. On the
other hand, the denotational (and derived non standard) semantics
strongly suggest operational semantics for the implementation of the
NXP architecture. Formally provable congruence relation henceforth
ascertain the soundness of the resulting implementation.

Finally the categorical foundation of the NXP architecture extends to
the formal representation of other cognitive processes beyond
inference. In particular the monadic style can be used
to articulate the working memory process to inference in a modular
way. It can also be used to express relationship between inference and
learning or adaptative processes reflecting the long term effect of experience
on inference.

\bibliographystyle{amsplain}
\bibliography{wip}
\end{document}